\documentclass{article}
\usepackage{spconf,amsmath,graphicx,hyperref, booktabs, multirow, array, caption, adjustbox, makecell, amssymb}

\title{SemiTooth: a Generalizable Semi-supervised Framework for Multi-Source Tooth Segmentation}
\name{
    Muyi Sun$^{1,\ast}$\thanks{$\ast$ These authors contributed equally to this work.}, 
    Yifan Gao$^{1,\ast}$, 
    Ziang Jia$^{2,3}$,  
    Xingqun Qi$^{4}$,
    Qianli Zhang$^{5}$, 
    Qian Liu$^{6}$, 
    Tianzheng Deng$^{6,\dagger}$\thanks{$\dagger$ Corresponding author.}%
    \thanks{This research was funded by the BNSF (No. QY25340) and the NSFC (No. 62306309).}
}
\address{
    $^{1}$School of AI, BUPT 
    $^{2}$CBDE 
    $^{3}$NLPR, CASIA 
    $^{4}$AIS, HKUST 
    $^{5}$PKU-SS 
    $^{6}$AFMC 
}

\begin{document}
\maketitle
\begin{abstract}
With the rapid advancement of artificial intelligence, intelligent dentistry for clinical diagnosis and treatment has become increasingly promising. 
As the primary clinical dentistry task, tooth structure segmentation for Cone-Beam Computed Tomography (CBCT) has made significant progress in recent years. 
However, challenges arise from the obtainment difficulty of full-annotated data, and the acquisition variability of multi-source data across different institutions, which have caused low-quality utilization, voxel-level inconsistency, and domain-specific disparity in CBCT slices.
Thus, the rational and efficient utilization of multi-source and unlabeled data represents a pivotal problem. 
In this paper, we propose \textbf{SemiTooth}, a \textbf{generalizable semi-supervised} framework for multi-source tooth segmentation. 
Specifically, we first compile \textbf{MS$^3$Toothset}, a \textbf{M}ulti-\textbf{S}ource \textbf{S}emi-\textbf{S}upervised Tooth DataSet for clinical dental CBCT, which contains data from three sources with different-level annotations.
Then, we design a multi-teacher and multi-student framework, i.e., SemiTooth, which promotes semi-supervised learning for multi-source data. 
SemiTooth employs distinct student networks that learn from unlabeled data with different sources, supervised by its respective teachers. 
Furthermore, a Stricter Weighted-Confidence Constraint is introduced for multiple teachers to improve the multi-source accuracy.
Extensive experiments are conducted on MS$^3$ Toothset to verify the feasibility and superiority of the SemiTooth framework, which achieves SOTA performance on the semi-supervised and multi-source tooth segmentation scenario.
\end{abstract}


\section{Introduction}
\label{sec:intro}
With the increasing attention to dental health, the demand for fast and efficient diagnosis has grown rapidly \cite{tooth_medical}. Tooth structure segmentation, a primary task in dental clinics, enhances diagnostic efficiency, accuracy, and stability \cite{ai4tooth, AI_Med_low_income}. Among imaging modalities, CBCT is widely used, making CBCT-based tooth segmentation a clinically important standard for orthodontics, implant planning, and lesion analysis.

In recent years, fully supervised CBCT methods have advanced clinical applications \cite{MIS_1, MIS_2, MIS_3}.
However, voxel-level annotations and large data requirements make the task labor-intensive and costly. 
Consequently, large amounts of de-identified and unlabeled CBCT data remain underutilized \cite{SSMIS_Survey}. 
To address these issues, semi-supervised medical image segmentation (SSMIS) methods leverage both labeled and unlabeled data to improve performance \cite{SSMIS_1, SSMIS_2, SSMIS_3}.
Nevertheless, most SSMIS approaches are design for single-source data \cite{SSDA_1, SSDA_2}, while dental CBCT from different institutions and devices exhibits substantial distribution gaps, limiting cross-source generalization \cite{DG}.In addition, open-source CBCT datasets for tooth segmentation remain scarce.

\graphicspath{{Images/}}
\begin{figure}[t]
\centering
\includegraphics[width=0.48\textwidth]{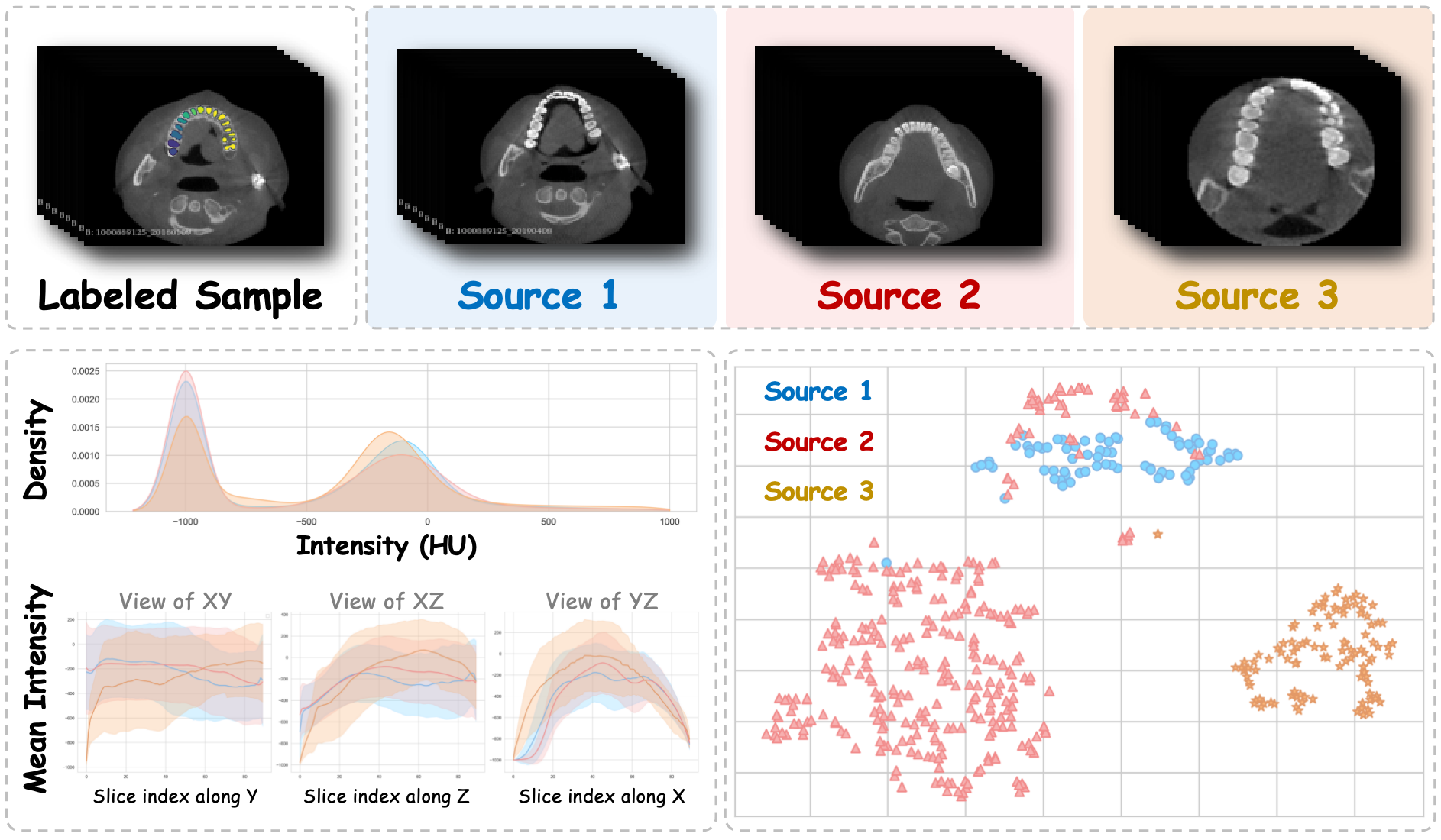}
\caption{Labeled and unlabeled samples of our Multi-Source Semi-Supervised \textbf{MS$^3$Toothset}. The comparisons of numerical (Density, Intensity) and feature distribution (t-SNE) illustrate the inherent source gaps. (Please zoom in for details.) }
\label{fig:dataset}
\vspace{-0.5cm}
\end{figure}

\graphicspath{{Images/}}
\begin{figure*}[t]
\centering
\includegraphics[width=0.98\textwidth]{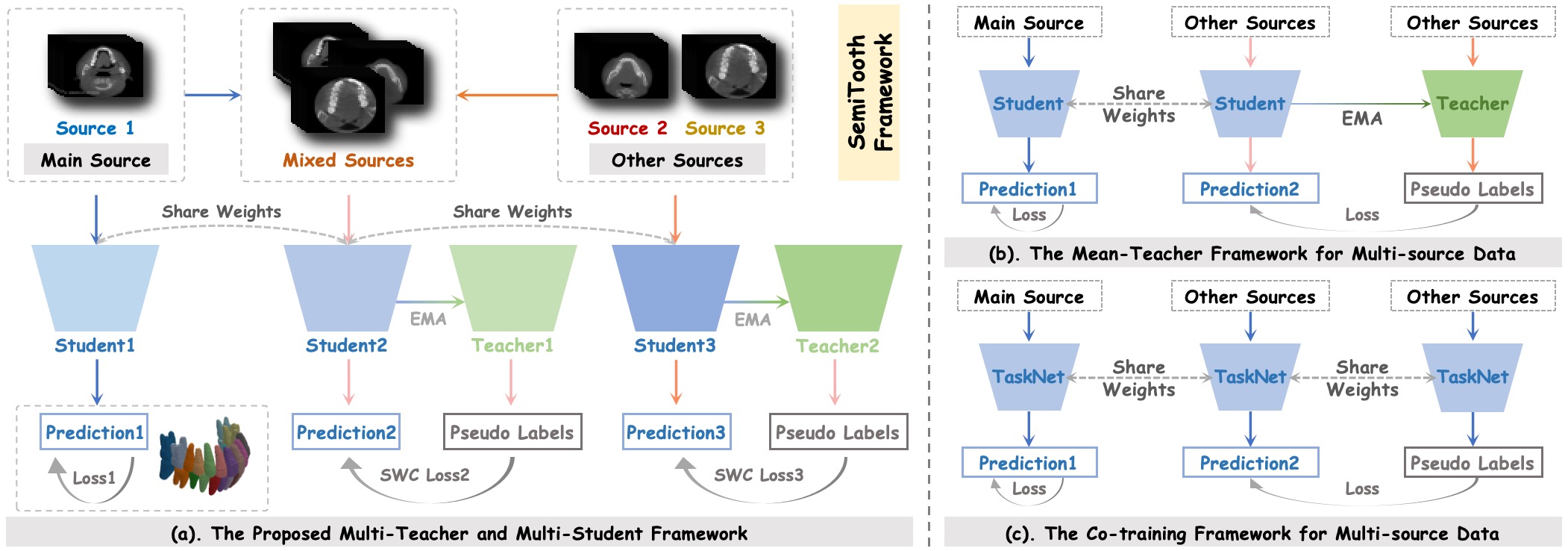}
\vspace{-0.2cm}
\caption{SemiTooth Framework.
(a) SemiTooth employs three students to handle different subsets, while two teachers supervise the mixed/other students to stabilize learning and improve pseudo-labels.  
(b) Mean Teacher is a basic framework that relies on a single teacher–student pair and lacks cross-source guidance.  
(c) Co-training uses multiple students (TaskNets) with shared weights but no teacher supervision, which fails to provide stable pseudo-labels.  
By combining multi-student collaboration and teacher guidance, SemiTooth effectively exploits multi-source data and enhances cross-source generalization.  %
}

\label{fig:SemiTooth}
\vspace{-0.5cm}
\end{figure*}

In summary, two main challenges remain for CBCT-based tooth segmentation. 
(1) Establishing a comprehensive multi-source CBCT dataset for clinical tooth analysis, 
(2) Developing a semi-supervised method capable of handling multi-source data. 
To address this, we first construct MS$^3$Toothset, a multi-source CBCT dataset collected from three sources. 
As shown in Fig.~\ref{fig:dataset}, the top row displays representative labeled and unlabeled samples, illustrating anatomical variations. 
Kernel Density Estimation curves \cite{MIS_3,kde}  and middle-slice intensity profiles in Hounsfield Units (lower-left) reveal voxel-level discrepancies, 
while t-SNE\cite{DG} embeddings (lower-right) show distinct feature clusters, highlighting substantial source gaps. 
These gaps challenge the unified model training across sources. 
To tackle this, we propose SemiTooth, a multi-branch semi-supervised framework where each teacher supervises its corresponding student to learn source-aware knowledge. 
We further introduce a region-level consistency constraint, the Stricter Weighted-Confidence Constraint for multiple teachers, which can refine pseudo-labels and enhance multi-source accuracy.

The main contributions are illustrated as follows:
(1) We compile MS$^3$Toothset, a Multi-Source Semi-Supervised Tooth DataSet for tooth structure segmentation.
(2) We propose SemiTooth, a generalizable semi-supervised framework, with multi-teacher and multi-student structure for multi-source tooth segmentation.
(3) A Stricter Weighted-Confidence Constraint is introduced for multiple teachers to improve the multi-source accuracy.
(4) Extensive experiments are conducted, where SemiTooth achieves SOTA on the semi-supervised multi-source tooth segmentation scenario.

\section{Related Work}
\label{sec:related}
\subsection{Semi-Supervised Medical Image Segmentation}
SSMIS effectively addresses the scarcity of annotated data \cite{SSMIS_Survey}, which is particularly relevant for CBCT tooth segmentation.  
The Teacher-Student paradigm, exemplified by Mean Teacher (MT) \cite{MT}, is a core methodology, with various extensions improving performance.  
For example, Zhang et al. \cite{SSMIS_1} leveraged correlations in unlabeled multi-modal data with soft pseudo-label re-learning. 
Wang et al. \cite{SSMIS_2} designed dual subnets exchanging knowledge via contrastive review and dynamic pseudo-labels. 
Su et al. \cite{SSMIS_3} proposed a dual-network mutual learning framework that selects reliable pseudo-labels via confidence and consistency.
However, these methods mainly focus on single-source data, limiting robustness across institutions, devices, or imaging modalities. 
This motivates the need for semi-supervised multi-source approaches capable of cross-source knowledge transfer.

\subsection{Multi-source Medical Image Analysis}
In clinical practice, single-source datasets are often small and variable in quality. Multi-source data are easier to obtain but exhibit substantial source gaps, degrading model performance \cite{DG}.  
To address this, Hu et al. \cite{MSMIS_1} forced models to learn generalized yet task-relevant features, and Wang et al. \cite{MSMIS_2} reduced domain gaps via pixel- and image-level uncertainty.  
These methods improve robustness but often require complex networks or strong supervision, limiting scalability with scarce labeled data.  
Semi-supervised strategies exploit unlabeled data alongside labeled sources. For example, Li et al. \cite{MSSSMIS_1} proposed a multi-branch framework where intra- and cross-source teachers guide a student, and Chen et al. \cite{MSSSMIS_2} used weak labels with self-disambiguation and hierarchical sampling.  
Despite these advances, publicly available multi-source CBCT tooth datasets remain scarce, and existing semi-supervised multi-source methods have seen limited validation.
This motivates our proposed generalizable semi-supervised framework for multi-source tooth segmentation.

\section{SemiTooth Framework}
\label{sec:SemiTooth}

\subsection{Overview}
In this section, we introduce \textbf{SemiTooth}, a generalizable framework for multi-source semi-supervised tooth segmentation.  
SemiTooth employs a multi-branch architecture to effectively leverage multi-source data.  
We also propose \textbf{Stricter Weighted-Confidence Constraint}, providing reliable guidance from multiple teachers during training.  
Together, this framework addresses data discrepancies across sources while enabling stable and accurate semi-supervised learning.

\subsection{SemiTooth}  
CBCT tooth segmentation suffers from scarce annotations and limited generalization across different sources. Existing SSMIS methods reduce annotation cost but struggle in multi-source settings with distribution gaps \cite{DG}. To address this, we propose \textbf{SemiTooth}, a multi-branch SSMIS framework for multi-source scenarios beyond Mean Teacher \cite{MT}.
In SemiTooth, all sources are reconstructed into three subsets: main (label), other (unlabel), and mixed. The mixed subset contains unlabeled samples that are distributionally similar to the main source, as measured by inter-source Wasserstein distance. This similarity bridges sources and improves training robustness.
As shown in Fig.~\ref{fig:SemiTooth}, SemiTooth employs three students and two teachers. Each student is assigned to one subset, ensuring that different sources are effectively utilized. The two teachers supervise the students on the mixed and other sources, to stabilize learning and improve pseudo-label quality. Students share similar architectures to promote effective knowledge transfer while maintaining sufficient diversity \cite{CMT, ShareWeights}. After each iteration, the corresponding teacher is updated via EMA of its student’s parameters:
\begin{equation}
\theta_t^{(k)} \leftarrow \gamma \, \theta_t^{(k-1)} + (1-\gamma)\, \theta_s^{(k)}
\end{equation}
where $\theta_t^{(k)}$ and $\theta_s^{(k)}$ denote the parameters of the $k$-th teacher and student, and $\gamma$ is the decay rate of EMA.  
Overall, SemiTooth achieves more robust learning and superior generalization compared with conventional semi-supervised frameworks on multi-source learning.

\subsection{Stricter Weighted-Confidence Constraint}
In SSMIS, CBCT heterogeneity introduces noise that degrades consistency regularization \cite{SSDA_1, SSDA_2}. To address this, we propose the \textbf{Stricter Weighted-Confidence} (SWC) constraint to extract reliable signals in multi-source learning.

\begin{figure}[t]
\centering
\includegraphics[width=0.485\textwidth]{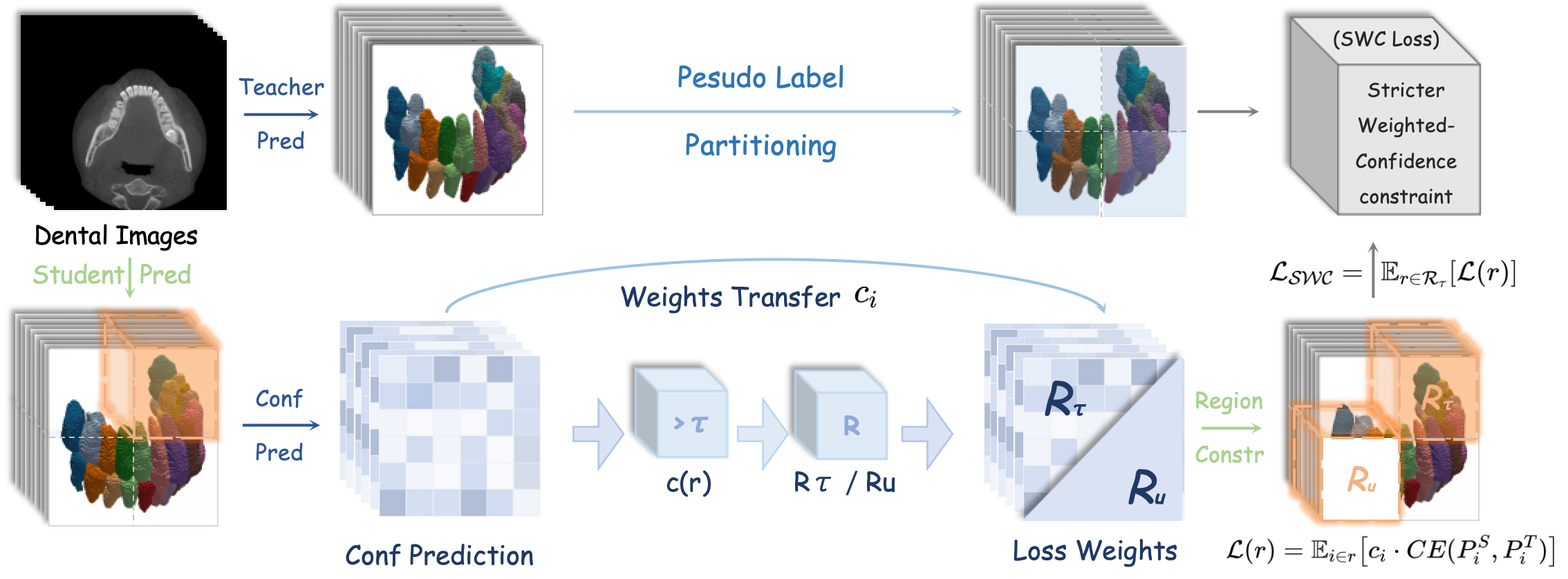}
\caption{The Stricter Weighted-Confidence Constraint. SWC adaptively emphasizes reliable regions while suppressing noisy predictions, ensuring cleaner consistency learning across multi-source CBCT data. (Please zoom in for details.)}
\label{fig:SWC}
\vspace{-0.6cm}
\end{figure}

After obtaining teacher and student probability distributions $P^T, P^S \in \mathbb{R}^{C\times D\times H\times W}$, each sample is evenly partitioned into non-overlapping cubic regions $\{r\}$. For each region $r$, the probability distribution is viewed as voxel-wise confidence, and the region confidence is defined as
\begin{equation}
c(r) = \mathbb{E}_{i \in r} \big[ \max_c P^T_{i,c} \big],
\end{equation}
where $\mathbb{E}_{i \in r}[\cdot]$ denotes the mean over all voxels $i$ in region $r$. Regions with $c(r) < \tau$ are considered unreliable and ignored, denoted as $\mathcal{R} _{u}$. While regions with $c(r) \ge \tau$ are retained, denoted as $\mathcal{R} _{\tau}$. Within each region in $\mathcal{R} _{\tau}$, voxel-wise confidence $c_i = \max_c P^T_{i,c}$ is preserved to weight the alignment between teacher and student outputs. The SWC constraint is then formally defined
\begin{equation}
\mathcal{SWC}(P^S,P^T) = \mathbb{E}_{r\in \mathcal{R} _{\tau}} \Big[ \mathbb{E}_{i \in r} \big[ c_i \cdot \mathcal{A}(P^S_i,P^T_i) \big] \Big],
\end{equation}
where $\mathbb{E}_{r: c(r)\ge\tau}[\cdot]$ denotes the mean over all valid regions, $\mathcal{A}(\cdot,\cdot)$ is a generic alignment operator, and $i$ indexes voxels within a region.
This formulation integrates region-level gating with voxel-wise weighting, thereby balancing structural reliability and voxel-level precision. 
An intuitive depiction of the constraint mechanism is provided in Fig.~\ref{fig:SWC}. 
The constraint is particularly well-suited for 3D CBCT tooth segmentation.

\graphicspath{{Images/}}

\subsection{Objective Functions} 
\textbf{Stricter Weighted-Confidence Loss.} 
Based on the SWC constraint, the loss for each region $r\in \mathcal{R} _{\tau}$ is defined as:
\begin{equation}
\mathcal{L}(r) = \mathbb{E}_{i \in r} \left[ c_i \cdot {CE}(P{i}^{S}, P_{i}^{T}) \right],
\end{equation}
and the overall SWC loss is given by:
\begin{equation}
\mathcal{L}_\mathcal{SWC}  = \mathbb{E}_{r \in \mathcal{R}_{\tau}} \left[ \mathcal{L}(r) \right],
\end{equation}
where $\text{CE}(\cdot,\cdot)$ denotes the cross-entropy loss.

\textbf{Semi-supervised Learning.} 
During training, each student receives two types of supervision. For labeled samples $x^l$ (main source), the supervised loss is
\begin{equation}
\mathcal{L}_{sup} = {CE}(P^S(x^l), y),
\end{equation}
where $y$ are the ground-truth labels.
For other sources and mixed sources, the SWC-based losses are computed:
\begin{equation}
\mathcal{L}_{cons}^{(u/h)} = \mathcal{L}_\mathcal{SWC}(P^S_i, P^T_i),
\end{equation}
where $u$ and $h$ denote different sources, respectively.

The total loss combines all components ,where $\alpha$ and $\beta$ balance the contributions of different sources.
\begin{equation}
\mathcal{L}_{total} = \mathcal{L}_{sup} + \alpha \, \mathcal{L}_{cons}^{u} + \beta \, \mathcal{L}_{cons}^{h},
\end{equation}

\graphicspath{{Images/}}
\begin{figure*}[t]
\centering
\includegraphics[width=0.95\textwidth]{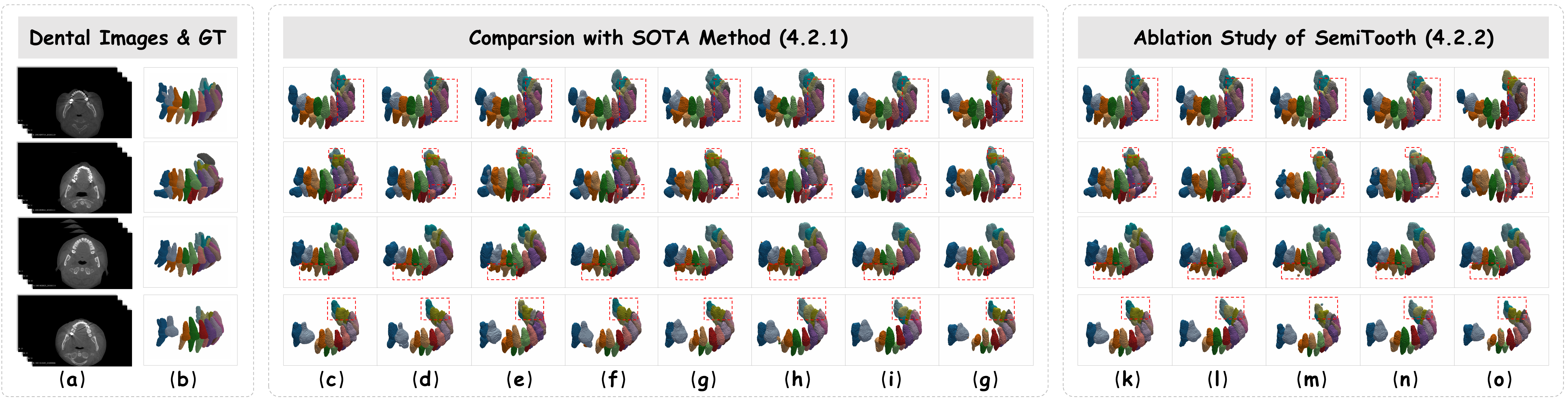}
\vspace{-0.2cm}
\caption{Comparison with different methods and Ablation Study on SemiTooth.
(a-b: Dental Images, Ground Truth). 
(c-g: SemiTooth, Uni-HSSL, CMT, MLRPL, ASDA, UA-MT, MT, V-Net). 
(k-o: Exp~5, ~4, ~3, ~2, ~1).
(\textbf{Please Zoom in for Details.)}
}
\label{fig:method_cmp}
\vspace{-0.5cm}
\end{figure*}


\section{Experimental Details and Results}
\vspace{-0.2cm}
\subsection{Dataset and Implementation details}
Experiments are conducted on the MS$^3$Toothset, which combines semi-annotated data from ShanghaiTech \cite{MIS_3} with proprietary unlabeled data from PKU-SS and AFMC. We have filtered and processed the data. Finally, this dataset contains 98 labeled samples (20 for testing) and 438 unlabeled samples. 
Experiments are conducted on four NVIDIA A4500 GPUs with batchsize of 4. V-Net \cite{vnet} is used as the backbone, and optimization is performed with Adam at a learning rate of 0.0001. The decay rate of EMA $\gamma$ is set to 0.99. In $\mathcal{L}_\mathcal{SWC}$, the threshold $\tau$ is set to 0.9, while in the total loss $\mathcal{L}_{total}$, the weights $\alpha$ and $\beta$ are both set to 0.5. Models are trained for 300 epochs. Segmentation performance is evaluated using mIoU and Dice. Recall is reported to ensure critical clinical sensitivity. Accuracy is used for pixel-wise correctness.

\vspace{-0.3cm}

\subsection{Experimental Results}
\subsubsection{Comparsion with SOTA Method}
Tab.~\ref{tab:comparison} shows the comparsions with different methods. contain generalizable single-source semi-supervised methods \cite{MT, ua-mt, SSMIS_3}.
Others are designed for multi-source semi-supervised learning \cite{SSDA_2, CMT, SSDA_3}. 
Our framework achieves better performance on both metrics in Tab.~\ref{tab:comparison} and result visualization in Fig.~\ref{fig:method_cmp} (c-g), demonstrating its effectiveness in leveraging unlabeled data from multi-source.

\vspace{-0.5cm}
\subsubsection{Ablation Study of SemiTooth}
To evaluate the components of SemiTooth, we conduct a series of ablation studies. The comparisons are shown in Tab.~\ref{tab:ablation} and Fig.~\ref{fig:method_cmp} (k-o). 
The baseline V-Net (Exp~1) shows limited performance. 
Mean Teacher (MT) (Exp~2) improves results, but still produces irregular tooth boundaries. 
Incorporating the SWC constraint (Exp~3) helps reduce noisy edges and enhances boundary sharpness.
Replacing Mean Teacher with SemiTooth (ST) (Exp~4) yields more natural tooth morphology, especially the root regions.
The full combination (Exp~5) produces the best results, creating more realistic tooth shapes with fewer boundary adhesions between neighbouring teeth.

\begin{table}[h]
\centering
\caption{Comparison with different methods.}
\vspace{-0.3cm}
\label{tab:comparison}
\adjustbox{max width=\linewidth,scale=0.75}{
\begin{tabular}{c|c|c|cccc}
\toprule
\textbf{Method} & \textbf{Publication} & \textbf{Year} & \textbf{mIoU} & \textbf{Dice} & \textbf{Recall} & \textbf{Acc} \\
\midrule
V-Net \cite{vnet}  & IEEE 3DV &  2016 & 61.36 & 73.65 & 70.77 & 66.75 \\
MT \cite{MT}       & NeurIPS &  2017 & 67.69 & 78.72 & 78.06 & 73.68 \\
UA-MT \cite{ua-mt} & MICCAI &  2019 & 68.37 & 79.18 & 80.42 & 76.17 \\
ASDA \cite{SSDA_2}   & IEEE TIP & 2022 & 73.75 & 83.63 & 80.93 & 78.79 \\     
MLRPL \cite{SSMIS_3}  & Elsevier MIA & 2024 & 72.86 & 83.29 & 79.75 & 77.39 \\    
CMT \cite{CMT}     & ACM MM & 2024 & 76.14 & 85.07 & 87.14 & 84.32 \\
Uni-HSSL \cite{SSDA_3}  & IEEE CVPR & 2025 & 75.76 & 85.42 & 84.26 & 81.88 \\
\textbf{Ours}  & \textbf{Draft(ICASSP)} & \textbf{2025} & \textbf{76.67} & \textbf{85.69} & \textbf{88.66} & \textbf{86.44} \\
\bottomrule
\end{tabular}
}
\end{table}
\vspace{-0.5cm}
\begin{table}[h]
\centering
\caption{Ablation Study of SemiTooth.}
\vspace{-0.3cm}
\label{tab:ablation}
\adjustbox{max width=0.74\linewidth}{ 
\begin{tabular}{c|cccc|cccc}
\toprule
\multirow{2}{*}{\textbf{Exp}} & \multicolumn{4}{c|}{\textbf{Modules}} & \multicolumn{4}{c}{\textbf{Metrics (\%)}} \\
\cmidrule(lr){2-5} \cmidrule(lr){6-9}
 & \makecell{V-Net} & \makecell{MT} & \makecell{ST} & \makecell{SWC} & \textbf{mIoU} & \textbf{Dice} & \textbf{Recall} & \textbf{Acc} \\
\midrule
1 & $\checkmark$ & & & & 61.36 & 73.65 & 70.77 & 66.75 \\
2 & $\checkmark$ & $\checkmark$ & & & 67.69 & 78.72 & 78.06 & 73.68 \\
3 & $\checkmark$ & $\checkmark$ & & $\checkmark$ & 69.94 & 80.29 & 79.67 & 75.34 \\
4 & $\checkmark$ & $\checkmark$ & $\checkmark$ & & 75.37 & 84.56 & 83.07 & 80.48 \\
5 & $\checkmark$ & $\checkmark$ & $\checkmark$ & $\checkmark$ & \textbf{76.67}  & \textbf{85.69}  & \textbf{88.66}  & \textbf{86.44} \\
\bottomrule
\end{tabular}}
\vspace{-0.4cm}
\end{table}

\subsubsection{Verification on Multi-source Data}
\begin{figure}[h]
\centering
\includegraphics[width=1\linewidth]{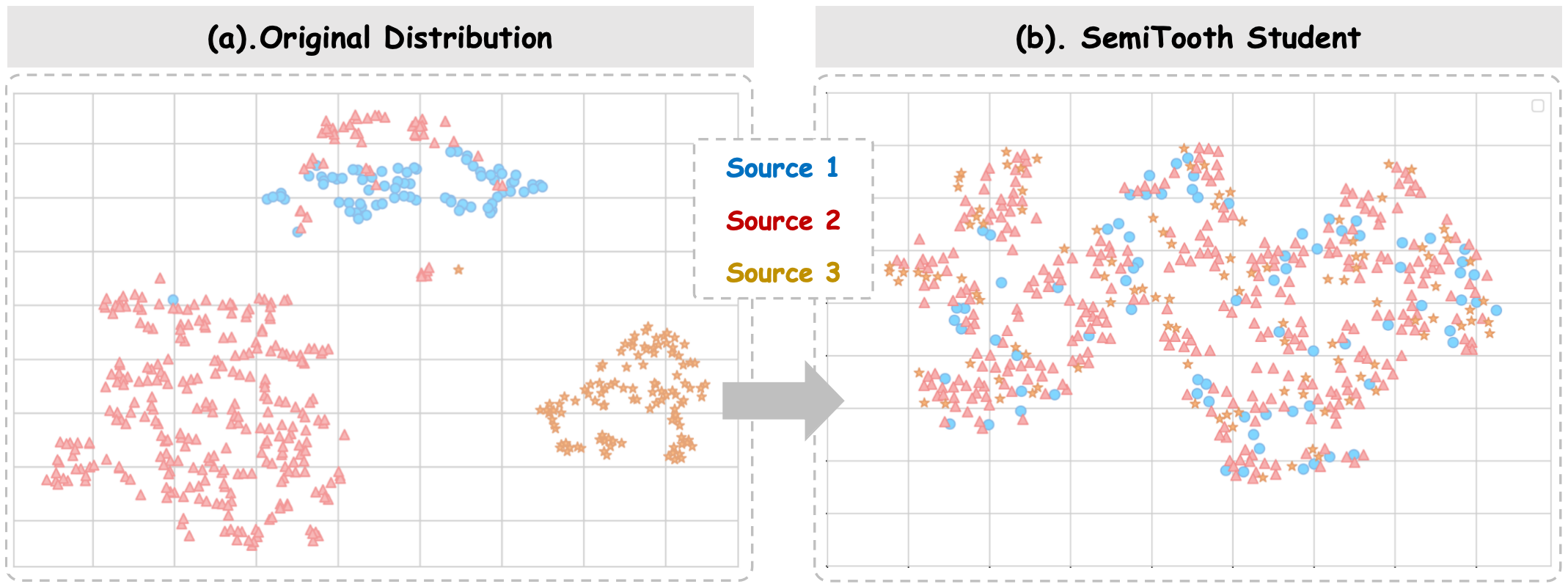}
\caption{Multi-source feature comparisons after SemiTooth.}
\label{fig:train_tsne}
\vspace{-0.6cm}
\end{figure}

To demonstrate the multi-source generalization, we visualize feature distributions (t-SNE) after the SemiTooth student. 
As shown in Fig~\ref{fig:train_tsne}, compared to the original three-soure data distributions (a), features from different sources become more clustered under our method (b), which demonstrate the generalization improvement and distribution gap narrowness.
\vspace{-0.3cm}

\graphicspath{{Images/}}

\vspace{-0.2cm}

\section{Conclusions}
\vspace{-0.2cm}
In this paper, we propose SemiTooth, a multi-branch semi-supervised framework for multi-source CBCT tooth segmentation.  
We also introduce the Stricter Weighted-Confidence Constraint to improve pseudo-label reliability and training stability.  
We construct the MS$^3$Toothset and show that SemiTooth achieves strong performance on it.  
Our research enables reliable tooth segmentation across multi-source CBCT, supporting practical diagnosis and treatment planning.

\vfill\pagebreak




\clearpage
\bibliographystyle{IEEEbib}
\bibliography{SemiTooth}

\end{document}